\documentclass[letterpaper]{article} 
\usepackage{aaai2027}
\usepackage[hyphens]{url} 
\usepackage{graphicx} 
\usepackage{natbib} 
\usepackage{caption} 
\usepackage{algorithm}
\usepackage{algorithmic}
\usepackage{booktabs}
\usepackage{amsmath}
\usepackage{amssymb}

\title{Beyond Uniform Compression: Budgeted Transmission Allocation for Extreme Federated Learning}
\author{
Pengfei Li,
Mohammad Khalil
}

\affiliations{
Centre for the Science of Learning \& Technology (SLATE),\\
University of Bergen, Bergen, Norway
}

\begin{document}

\maketitle

\begin{abstract}
Federated learning faces severe communication bottlenecks when clients upload high-dimensional model updates. Existing methods often compress these updates uniformly across all layers. This uniform approach ignores the heterogeneous value of different parameter blocks and wastes limited bandwidth on insensitive layers. To address this issue, we propose Layer-wise Budgeted Adaptive Transmission (LBAT). LBAT reframes federated communication under extreme uplink budgets as a resource allocation problem. Our framework dynamically estimates the transmission value of different layers utilising local training signals. It then employs an exact byte dynamic programming allocator to determine optimal rank and bit configurations under strict budgets. We validate LBAT on highly heterogeneous federated tabular prediction and data generation tasks. Extensive experiments demonstrate that LBAT consistently outperforms uniform rank, uniform quantisation, and fixed compression baselines across various extreme budget regimes. Furthermore, it achieves significantly better communication and utility tradeoffs while preserving essential distributional fidelity.
\end{abstract}

\section{Introduction}

Federated learning enables collaborative model training without centralising raw data, offering a vital paradigm for domains like healthcare and finance that face strict egress and compliance constraints \cite{mcmahan2017communication,kairouz2021advances,yang2019federated}. The classic Federated Averaging (FedAvg) framework enables data-localised model training via multiple-step client local updates and periodic server aggregation, becoming the foundational paradigm for subsequent federated optimisation methods \cite{mcmahan2017communication,kairouz2021advances}. However, FedAvg fails to eliminate federated learning system bottlenecks: clients must still upload high-dimensional model updates to servers each communication round. As client numbers increase, model scales expand, or uplinks face constraints, repeated model update uploads rapidly dominate training costs and deployment feasibility \cite{konevcny2016federated,bonawitz2019towards}. Thus, communication-efficient learning has long focused on message compression, synchronisation frequency control, and partial client participation. Representative methods include QSGD, balancing communication volume and convergence error via stochastic quantisation \cite{alistarh2017qsgd}, Deep Gradient Compression, significantly reducing distributed training bandwidth overhead via gradient sparsification \cite{lin2017deep}, and FedPAQ, combining periodic averaging, partial client participation, and quantised message passing \cite{reisizadeh2020fedpaq}. Recently, parameter-efficient fine-tuning provides another path to reduce federated communication costs: LoRA significantly cuts trainable parameters by freezing backbone models and training only low rank adapters \cite{hu2022lora}, while subsequent federated LoRA methods explore uploading and aggregating only lightweight adapters, or incorporating quantisation to compress client updates \cite{babakniya2023slora,grativol2024flocora}.

However, most aforementioned communication-efficient methods focus on encoding given model updates more compactly, reducing single-round communication overhead via quantisation, sparsification, low-rank approximation, or lowered synchronisation frequency \cite{konevcny2016federated,alistarh2017qsgd,lin2017deep,reisizadeh2020fedpaq,albasyoni2020optimal}. When per-round upload budgets are extremely small, merely compressing full updates is insufficient. Clients must also decide which layers, adapter matrices, and parameter blocks merit uploading most. Existing uniform quantisation, uniform rank allocation, Top $k$ sparsification, or fixed LoRA configurations typically assume different parameter blocks possess approximately equal communication value \cite{hu2022lora,zhang2023adalora}. But deep model hierarchical structures exhibit significant heterogeneity: different layers vary in parameter scale, compression sensitivity, and task utility contribution. Multiple adaptive optimisation studies show layer-wise or matrix-wise budget allocation is often more effective than uniform configuration \cite{dong2020hawq,zhang2023adalora,markov2024greco}. In federated learning, this issue is further exacerbated by client data heterogeneity and aggregation shift. Non-IID data induces client drift, and different layers may exhibit varying model differences across clients \cite{karimireddy2020scaffold,li2020fedprox,lee2023layer}. Thus, in heterogeneous client scenarios with extremely constrained uplink budgets, key problems include not only "how to compress updates" but also "how to locally estimate parameter block values per client and select the most valuable adapter configuration for upload under strict byte budgets".

\begin{figure*}[t]
    \centering
    \includegraphics[width=\linewidth]{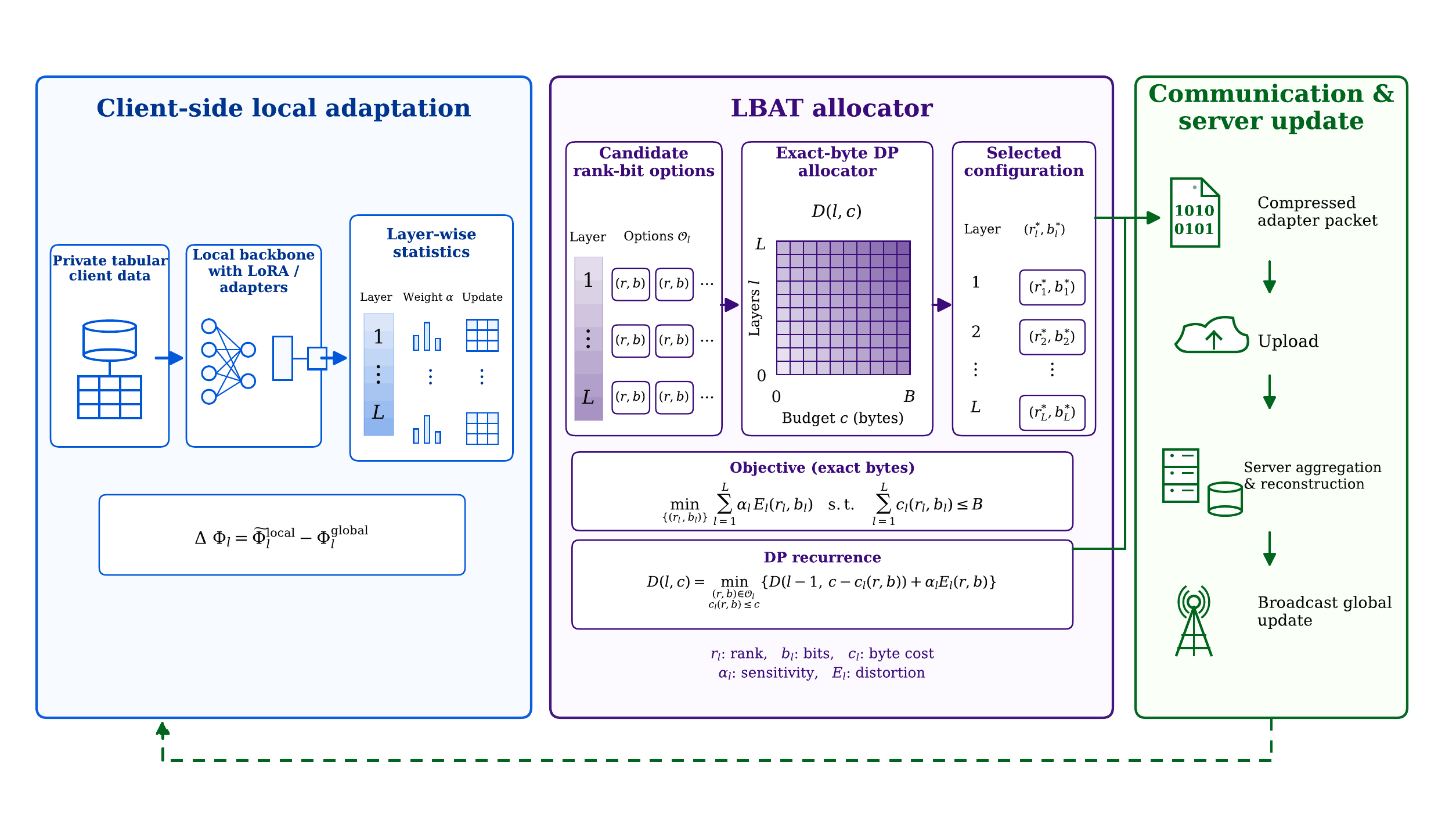}
    \caption{Overview of the Layer-wise Budgeted Adaptive Transmission (LBAT) framework. LBAT evaluates layer-wise sensitivity locally and employs an exact-byte dynamic programming allocator to determine the optimal rank and bit configuration $(r_l^*, b_l^*)$ under a strict communication budget $B$.}
    \label{fig:lbat_overview}
\end{figure*}

To this end, we propose LBAT (Layer-wise Budgeted Adaptive Transmission), a layer-wise adaptive transmission framework for extreme uplink communication budgets. LBAT's core perspective is: when communication is extremely restricted, the key problem lies not only in "how to compress model updates", but also in "which parameter blocks limited upload bytes should be preferentially allocated to". LBAT dynamically estimates the transmission values of different layers based on local training signals, and adaptively solves layer-wise upload configurations under strict budgets, thereby transforming federated communication from a "uniform compression" paradigm into a "budgeted value allocation" problem. The overall pipeline of our proposed LBAT framework is illustrated in Figure \ref{fig:lbat_overview}.

We validate LBAT in two highly challenging federated tabular learning scenarios: communication-constrained tabular prediction tasks (FedLBAT-Pred) and tabular data generation tasks (FedLBAT-Syn). These two scenarios cover dual prediction and generation objectives, and encompass realistic challenges like highly non-IID data, mixed features, and extreme sample skew. Extensive experiments show that LBAT stably outperforms baseline methods like uniform rank, uniform quantisation, FLoCoRA-style compression, and sensitivity-free allocation, achieving significantly better communication-utility trade-offs under extreme communication budgets. Our main contributions are as follows:

\begin{itemize}
    \item \textbf{Problem Reframing:} We reframe federated communication under extreme uplink budgets as a layer-wise transmission allocation problem, driving communication-efficient federated learning to shift from the "how much to compress" perspective to the "what to transmit preferentially" perspective.
    \item \textbf{Algorithm Design:} We propose LBAT, a lightweight framework not relying on public data, extra generators, or server-side distillation, maximising transmission marginal utility under strict budgets through local value estimation and adaptive allocation.
    \item \textbf{Empirical Validation \& Diagnostics:} We demonstrate LBAT's effectiveness in federated tabular prediction and generation tasks (FedLBAT-Pred \& FedLBAT-Syn). Additionally, through deep analysis of layer-wise allocation strategies, gap-to-full-upload, TRTR-normalized utility, and class coverage, we reveal the internal mechanisms and interpretability of LBAT's performance gains.
\end{itemize}

\section{Related Work}

\section{Communication Efficient Federated Learning}
Communication overhead remains a federated learning bottleneck. FedAvg significantly reduces communication rounds via local multi-step updates and server-side model averaging, but clients must still upload model updates per round, and uplink bandwidth often remains the primary limitation in cross-institution collaboration and privacy-sensitive tabular data scenarios \cite{mcmahan2017communication}. Therefore, extensive work studies compressing client upload content. QSGD establishes trade-offs between communication bits and optimisation variance via stochastic gradient quantisation \cite{alistarh2017qsgd}, Deep Gradient Compression reduces communication bandwidth utilising gradient sparsity \cite{lin2017deep}, and FedPAQ further lowers federated training communication costs combining periodic averaging and quantised communication \cite{reisizadeh2020fedpaq}. These methods primarily focus on "how to compress given updates", like reducing gradient precision or sparsifying updates. A related but finer-grained work category studies layer-wise compression. L GreCo notes different network layers exhibit varying compression error sensitivities, improving compression efficiency while satisfying error constraints via layer-wise adaptive compression parameter selection \cite{markov2024greco}. This paper studies budgeted transmission allocation in federated uploads, namely deciding which layers or parameter blocks merit transmission under extreme uplink budgets, rather than merely applying uniform or layer-wise compression to existing gradient updates.

\section{Parameter Efficient Updates in Federated Learning}
Parameter-efficient fine-tuning provides another approach for communication-constrained federated learning. LoRA significantly reduces trainable parameters by freezing main models and injecting low-rank trainable matrices \cite{hu2022lora}, a mechanism initially used for large model fine-tuning but naturally suitable as lightweight update carriers in federated communication. FLoCoRA introduces LoRA into federated training and further reduces communication costs by combining quantisation, showing low-rank adapters serve as strong communication compression baselines \cite{grativol2024flocora}. Recent works also study federated fine-tuning with heterogeneous LoRA rank, for example, HetLoRA allows different clients to use different ranks and designs corresponding aggregation strategies \cite{cho2024heterogeneous}. Such methods typically treat adapter structures or rank as part of model design, while this paper treats low-rank adapters as selectable and encodable transmission units, focusing on jointly allocating rank, bit width, and upload decisions across different layers given single-round uplink budgets.

\section{Federated Tabular Learning and Tabular Modeling}
Recently, diffusion models have been applied to tabular modelling: TabDDPM proves that diffusion models can handle continuous and discrete mixed-type tabular data, outperforming several GAN/VAE alternatives on multiple tabular benchmarks \cite{kotelnikov2023tabddpm}. In federated scenarios, FedTabDiff extends tabular diffusion models to multiple client data owners, enabling collaborative mixed-type tabular generative model training without centralising raw data \cite{sattarov2024fedtabdiff}, and DP FedTabDiff further combines differential privacy with federated tabular diffusion \cite{sattarov2025federated}. These works focus on tabular generative model construction and training, whereas this paper focuses on communication allocation problems in federated tabular learning and synthesis: how to allocate transmission resources across different layers or parameter blocks given byte budgets, thereby improving utility retention capability under communication-constrained conditions.

\section{Methodology}

This paper asks: when clients can upload extremely few bytes per round, which layers or parameter blocks merit transmission the most? LBAT models client uploads as a layer-wise budget allocation problem. Each transmittable parameter block possesses several candidate transmission configurations. Clients select configurations that minimise estimated losses within given byte budgets, based on local parameter sensitivities and communication costs. The server uses no distillation or extra proxy data, executing only sample-size-weighted aggregation on the received parameter updates.

\subsection{Budgeted Federated Transmission}\label{subsec:problem}

Consider federated learning involving $N$ clients. Let the global transmittable parameter blocks be denoted as
\begin{equation}
\boldsymbol{\Phi}=\{\Phi_1,\Phi_2,\ldots,\Phi_L\},
\end{equation}
where $L$ denotes the number of layers or parameter blocks participating in LBAT decisions. In round $t$, the server broadcasts the current parameters $\boldsymbol{\Phi}^{t}$. Client $i$ trains on its local data $\mathcal{D}_i$ to obtain local parameters $\widetilde{\boldsymbol{\Phi}}_i^{t}$. The corresponding block-level updates are
\begin{equation}
\Delta_{i,\ell}^{t}=\widetilde{\Phi}_{i,\ell}^{t}-\Phi_{\ell}^{t},\qquad \ell=1,\ldots,L.
\end{equation}
In parameter-efficient adapter models, $\Delta_{i,\ell}^{t}$ represents the equivalent weight updates generated by the $\ell$-th adapter block. If the model also contains a few auxiliary parameters that do not participate in LBAT decisions but require synchronisation (e.g., categorical embeddings or decoding heads), we treat them as ordinary federated parameters and deduct their communication volume from the current round's total budget. Thus, LBAT acts strictly on the remaining effective budget $\bar B_i^t$.

For each block $\ell$, the client constructs a finite candidate set $\mathcal{O}_{i,\ell}^{t}$. A candidate $o\in\mathcal{O}_{i,\ell}^{t}$ represents a specific transmission configuration, defined by hyperparameters such as the retained rank $r$ and the quantisation bit-width $b$. We denote the communication cost of this candidate in bytes as $C_{i,\ell}(o)$, and the server-decoded approximate update as $\widehat{\Delta}_{i,\ell}^{t}(o)$. We measure the candidate transmission distortion using the normalised reconstruction error:
\begin{equation}
E_{i,\ell}^{t}(o)=\frac{\left\|\Delta_{i,\ell}^{t}-\widehat{\Delta}_{i,\ell}^{t}(o)\right\|_{\mathrm{F}}^{2}}{\left\|\Delta_{i,\ell}^{t}\right\|_{\mathrm{F}}^{2}+\varepsilon},
\label{eq:distortion}
\end{equation}
where $\varepsilon>0$ is a numerical stability term to prevent division by zero. If a candidate $o$ signifies not uploading this block at all, then $C_{i,\ell}(o)=0$ and $\widehat{\Delta}_{i,\ell}^{t}(o)=0$.

\subsection{Sensitivity-Aware Objective}\label{subsec:sensitivity}

Different parameter blocks impact the final utility differently. Therefore, LBAT maintains a sensitivity score $\alpha_{i,\ell}^{t}$ for each block to measure the relative cost of its compression error. This weighting is motivated by block-level perturbation analysis. Let the compression error be $\delta_{i,\ell}^{t}=\widehat{\Delta}_{i,\ell}^{t}(o)-\Delta_{i,\ell}^{t}$. Assuming the objective function $\mathcal{F}$ satisfies block-wise smoothness, for a small perturbation $\delta_\ell$, we have:
\begin{equation}
\mathcal{F}(\boldsymbol{\Phi}+\delta_{\ell})\le\mathcal{F}(\boldsymbol{\Phi})+\left\langle \nabla_{\Phi_{\ell}}\mathcal{F}(\boldsymbol{\Phi}),\delta_{\ell}\right\rangle+\frac{L_{\ell}}{2}\left\|\delta_{\ell}\right\|_{\mathrm{F}}^{2}.
\end{equation}
When the model approaches a local minimum, or assuming the compression error direction is approximately orthogonal to the gradient, the impact of the first-order term is diminished. The utility drop caused by compression is then heuristically bounded by the second-order term $\frac{L_\ell}{2}\|\delta_\ell\|_{\mathrm{F}}^2$.

To approximate $L_\ell$ or the block-wise importance without incurring extra backward passes, LBAT reuses the gradient statistics naturally generated during local training. The sensitivity is updated online via an exponential moving average:
\begin{equation}
\alpha_{i,\ell}^{t}\leftarrow\rho\alpha_{i,\ell}^{t-1}+(1-\rho)\left\|g_{i,\ell}^{t}\right\|_{\mathrm{F}}^{2},
\label{eq:alpha}
\end{equation}
where $\rho\in[0,1)$, and $g_{i,\ell}^{t}$ is the accumulated gradient statistic of block $\ell$ during the round $t$ local training. 

Given the candidate sets, costs, distortions, and sensitivities, client $i$'s upload decision is formulated as:
\begin{equation}
\begin{aligned}
\min_{o_{i,1},\ldots,o_{i,L}} \quad& \sum_{\ell=1}^{L}\alpha_{i,\ell}^{t} E_{i,\ell}^{t}(o_{i,\ell}) \\
\text{s.t.}\quad& \sum_{\ell=1}^{L} C_{i,\ell}(o_{i,\ell}) \le \bar B_i^t, \\
& o_{i,\ell}\in\mathcal{O}_{i,\ell}^{t},\quad \ell=1,\ldots,L .
\end{aligned}
\label{eq:lbat_objective}
\end{equation}
Unlike uniform rank or uniform quantisation baselines that rigidly apply identical configurations to all layers, Equation~\eqref{eq:lbat_objective} allows LBAT to adaptively allocate the byte budget based on layer-wise marginal values.

\subsection{Solving the Allocation via Dynamic Programming}\label{subsec:lbat_solver}

Since each parameter block must select exactly one configuration from a discrete candidate set, Equation~\eqref{eq:lbat_objective} forms a Multiple-Choice Knapsack Problem (MCKP). Let $D[\ell,c]$ denote the minimum weighted distortion when processing the first $\ell$ parameter blocks using no more than $c$ bytes. The exact-byte dynamic programming recurrence is:
\begin{equation}
D[\ell,c]=\min_{\substack{o\in\mathcal{O}_{i,\ell}^{t} \\ C_{i,\ell}(o)\le c}}\left\{D[\ell-1,c-C_{i,\ell}(o)]+\alpha_{i,\ell}^{t}E_{i,\ell}^{t}(o)\right\} .
\label{eq:dp}
\end{equation}
We initialize $D[0,c]=0$. Finally, we select the optimal state satisfying $c\le\bar B_i^t$ from $D[L,c]$, obtaining the optimal layer-wise configurations via backtracking. The overall procedure is summarised in Algorithm~\ref{alg:fedlbat}.

\begin{algorithm}[t]
\caption{FedLBAT: Layer wise Budgeted Adaptive Transmission}\label{alg:fedlbat}
\begin{algorithmic}[1]
\REQUIRE Initial parameters $\boldsymbol{\Phi}^{0}$, client set $\{1,\ldots,N\}$, rounds $T$, effective byte budget per client $\bar B_i^t$.
\FOR{$t=0,1,\ldots,T-1$}
\STATE Server samples client subset $\mathcal{S}_t$ and broadcasts $\boldsymbol{\Phi}^{t}$.
\FOR{each $i\in\mathcal{S}_t$ in parallel}
\STATE Client $i$ trains locally, obtaining block updates $\{\Delta_{i,\ell}^{t}\}_{\ell=1}^{L}$ and sensitivities $\{\alpha_{i,\ell}^{t}\}_{\ell=1}^{L}$.
\STATE Construct candidate sets $\{\mathcal{O}_{i,\ell}^{t}\}_{\ell=1}^{L}$, evaluating cost $C_{i,\ell}(o)$ and distortion $E_{i,\ell}^{t}(o)$ per candidate.
\STATE Solve Eq.~\eqref{eq:lbat_objective} via exact byte DP (Eq.~\eqref{eq:dp}) to find optimal configurations $\{o_{i,\ell}^{t,*}\}_{\ell=1}^{L}$.
\STATE Upload compressed updates $\{\widehat{\Delta}_{i,\ell}^{t}(o_{i,\ell}^{t,*})\}_{\ell=1}^{L}$.
\ENDFOR
\STATE Server decodes updates and aggregates by sample size:
\STATE \hspace{1em}$\displaystyle \boldsymbol{\Phi}^{t+1}\leftarrow\boldsymbol{\Phi}^{t}+\frac{1}{\sum_{j\in\mathcal{S}_t}n_j}\sum_{i\in\mathcal{S}_t}n_i\widehat{\boldsymbol{\Delta}}_{i}^{t}$.
\ENDFOR
\RETURN $\boldsymbol{\Phi}^{T}$
\end{algorithmic}
\end{algorithm}

\subsection{Task Instances}\label{subsec:task_instances}
LBAT is an agnostic transmission-layer algorithm, independent of the underlying network architecture. We instantiate it across two distinct tabular tasks.

\paragraph{FedLBAT-Syn.} This task applies LBAT to communication-constrained federated tabular modelling. Clients locally train a tabular diffusion model where continuous and categorical features are encoded into a unified latent space $z_0$. The local denoising objective is:
\begin{equation}
\mathcal{L}_{\mathrm{syn}}=\mathbb{E}_{t, \epsilon \sim \mathcal{N}(0, I)}\left\|\epsilon-\epsilon_\theta(z_t,t)\right\|_2^2+\lambda\mathcal{L}_{\mathrm{cat}},
\end{equation}
where $\epsilon_\theta$ is the noise-prediction network and $\mathcal{L}_{\mathrm{cat}}$ is the categorical reconstruction loss. LBAT acts exclusively on the model updates during transmission without altering this specific generative loss. 

\paragraph{FedLBAT-Pred.} To verify that LBAT is not tied strictly to generative evaluators, we also apply it to standard tabular prediction. For binary classification, clients optimise the standard Binary Cross-Entropy (BCE) loss:
\begin{equation}
\begin{aligned}
\mathcal{L}_{\mathrm{pred}} = -\frac{1}{|\mathcal{B}|}\sum_{(x,y)\in\mathcal{B}} &\Big[ y\log \sigma(h_{\theta}(x)) \\
&+ (1-y)\log\left(1-\sigma(h_{\theta}(x))\right) \Big].
\end{aligned}
\end{equation}
Following local training, LBAT applies the identical budgeted transmission selection to the predictive model's parameter updates.

\section{Experimental Setup}

\subsection{Datasets and Federated Scenarios}
We evaluate LBAT on three binary tabular benchmarks covering demographic, financial, and medical domains: Adult\cite{kohavi1996adult}, Credit Default\cite{yeh2009comparisons}, and Cardio\cite{sulianova2019cardiovascular}.
We reserve 20\% of each dataset for testing and distribute the rest among $N=20$ clients using a Dirichlet label skew partition with $\alpha=0.3$ combined with quantity skew.
This creates severe sample size and label heterogeneity.
Each communication round samples 4 clients.
We deploy LBAT in two distinct settings.
The primary setting \textsc{Syn} trains a federated tabular diffusion model based on TabDDPM and FedTabDiff \cite{kotelnikov2023tabddpm,sattarov2024fedtabdiff} and evaluates generated tables via the Train on Synthetic Test on Real (TSTR) protocol.
The auxiliary setting \textsc{Real} applies the same budgeted transmission policy to a tabular MLP predictor.
This confirms our framework is broadly applicable to standard predictive models rather than just synthetic data evaluators.

\subsection{Baselines and Evaluation Protocol}
We define the $1.00\times$ full update budget as transmitting all adapter blocks at maximum rank and highest precision alongside synchronised auxiliary parameters.
Our experiments test five relative upload budgets ranging from $0.05\times$ to $1.00\times$.
We compare FedLBAT against five baseline methods.
\textsc{Uniform-Rank} assigns the same retained rank to all layers.
\textsc{Uniform-Quant} applies a uniform bit width across layers.
\textsc{FLoCORA-style} implements a fixed parameter efficient compression strategy \cite{hu2022lora,grativol2024flocora}.
\textsc{FedLAMA} adaptively adjusts the temporal aggregation intervals of different layers to reduce communication \cite{lee2023layer}.
\textsc{FedAvg-Full} transmits the complete adapter update \cite{mcmahan2017communication}.
We also include centralised and Train on Real Test on Real (TRTR) oracles as upper bounds.
For the synthetic track, our primary metric is TRTR normalised utility computed as $(\mathrm{AUC}_{\mathrm{TSTR}}-0.5) / (\mathrm{AUC}_{\mathrm{TRTR}}-0.5)$.
We additionally evaluate marginal Wasserstein distance correlation error label gap and categorical support.
The real prediction track reports test AUC and F1.
All main results are averaged over three random seeds.

\section{Results}

We organise the results around four questions:
(i) whether FedLBAT improves the communication--utility frontier under extreme upload budgets;
(ii) whether its advantage is consistent across datasets and budget regimes;
(iii) what layer-wise allocation policy LBAT learns;
and (iv) whether the same transmission principle transfers to real tabular prediction.
Unless otherwise stated, we report TRTR-normalized utility for the synthetic track, so that utility scores are comparable across datasets with different real-data predictability.

\subsection{Communication and Utility Tradeoffs Under Extreme Budgets}

\begin{figure}[t]
    \centering
    \includegraphics[width=\columnwidth]{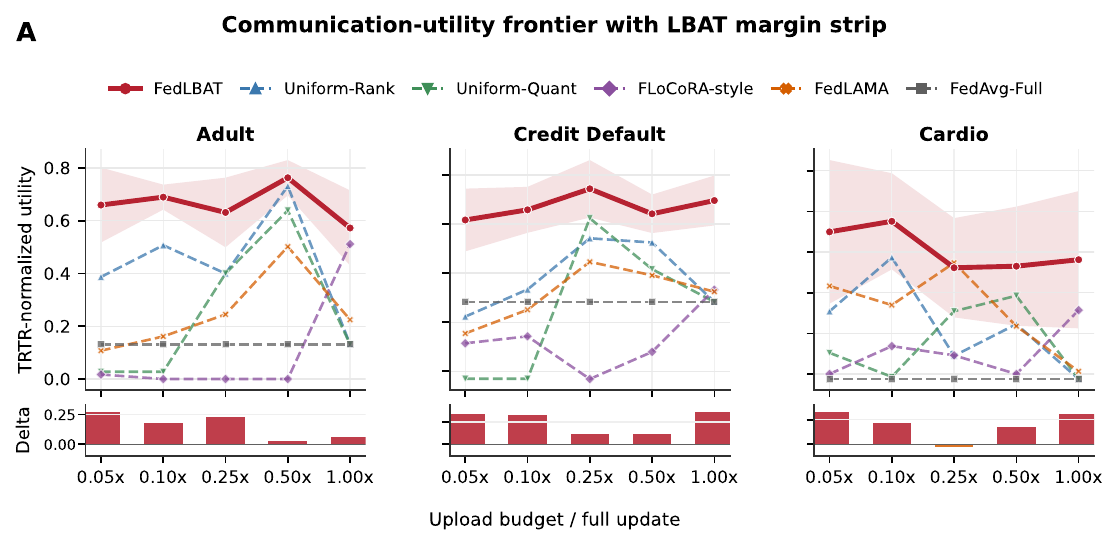}
    \caption{
    \textbf{Communication--utility frontier.}
    The horizontal axis is upload budget normalised by full update, and the vertical axis is TRTR-normalized utility.
    The margin strip below each panel shows the average utility gain of FedLBAT over the strongest communication-constrained peer.
    FedLBAT achieves a stronger frontier across all three datasets, particularly in the $0.05\times$--$0.25\times$ extreme-budget range.
    }
    \label{fig:comm_utility_frontier}
\end{figure}

Figure~\ref{fig:comm_utility_frontier} shows the communication--utility frontier under five upload budgets from $0.05\times$ to $1.00\times$ full update.
FedLBAT consistently forms a better frontier than uniform and FLoCORA-style baselines across Adult, Credit Default, and Cardio.
The advantage is most pronounced in the extreme-budget interval from $0.05\times$ to $0.25\times$ full update.
This supports the central claim of this paper: when only a small fraction of the full update can be transmitted, uniformly assigning the same rank or bit width to all layers wastes communication budget, while layer-wise budget allocation preserves more task-relevant information.

\subsection{Consistency Across Budgets and Datasets}

Table~\ref{tab:budget_dominance_detail} provides a detailed budget dataset view of FedLBAT dominance. 
Each margin is computed as FedLBAT minus the strongest non-FedLBAT communication-constrained peer under the same dataset and budget. 
FedLBAT obtains positive margins in $14$ out of $15$ dataset budget mean cells. 
The largest average margins appear at $0.05\times$ and $0.10\times$ where the upload budget is most restrictive. 
Across seed-level comparisons, FedLBAT achieves $37$ wins $5$ ties and $3$ losses. 
These results indicate that FedLBAT's advantage is not caused by a single favourable dataset or budget point but instead appears consistently across the full budget sweep.

\begin{table*}[t]
\centering
{%
\fontsize{9pt}{10pt}\selectfont
\setlength{\tabcolsep}{1mm}
\begin{tabular}{@{}lccccccc@{}}
\toprule
Budget
& \shortstack{Adult\\$\Delta$}
& \shortstack{Adult\\W/T/L}
& \shortstack{Credit\\$\Delta$}
& \shortstack{Credit\\W/T/L}
& \shortstack{Cardio\\$\Delta$}
& \shortstack{Cardio\\W/T/L}
& \shortstack{Mean\\$\Delta$} \\
\midrule
$0.05\times$
& +0.273 (+17.1\%) & 3/0/0
& +0.334 (+15.8\%) & 3/0/0
& +0.133 (+6.9\%) & 2/0/1
& +0.247 (+13.3\%) \\

$0.10\times$
& +0.182 (+10.6\%) & 3/0/0
& +0.325 (+15.0\%) & 3/0/0
& +0.089 (+4.4\%) & 3/0/0
& +0.199 (+10.0\%) \\

$0.25\times$
& +0.231 (+14.4\%) & 3/0/0
& +0.120 (+4.9\%) & 3/0/0
& -0.012 (-0.6\%) & 2/0/1
& +0.113 (+6.2\%) \\

$0.50\times$
& +0.032 (+1.6\%) & 0/3/0
& +0.117 (+4.9\%) & 3/0/0
& +0.073 (+3.8\%) & 2/0/1
& +0.074 (+3.5\%) \\

$1.00\times$
& +0.061 (+3.5\%) & 3/0/0
& +0.363 (+16.8\%) & 2/1/0
& +0.124 (+6.6\%) & 2/1/0
& +0.183 (+9.0\%) \\
\midrule
All
& -- & 12/3/0
& -- & 14/1/0
& -- & 11/1/3
& +0.163 (+8.4\%) \\
\bottomrule
\end{tabular}
}

\caption{
Detailed dominance summary across all upload budgets.
For each dataset--budget cell, $\Delta$ is FedLBAT minus the strongest non-FedLBAT peer in TRTR-normalized utility; parenthesised values report the relative TSTR-AUC gain,
$100(\overline{A}_{\mathrm{FedLBAT}}-\overline{A}_{\mathrm{peer}})
 /\overline{A}_{\mathrm{peer}}$, computed from unrounded three-seed mean AUCs.
W/T/L reselects the strongest peer within each matched seed
($|\Delta_s|\le 10^{-6}$ is a tie), while Mean percentages macro-average the corresponding cellwise AUC gains.
}
\label{tab:budget_dominance_detail}
\end{table*}

\begin{figure}[t]
    \centering
    \includegraphics[width=\columnwidth]{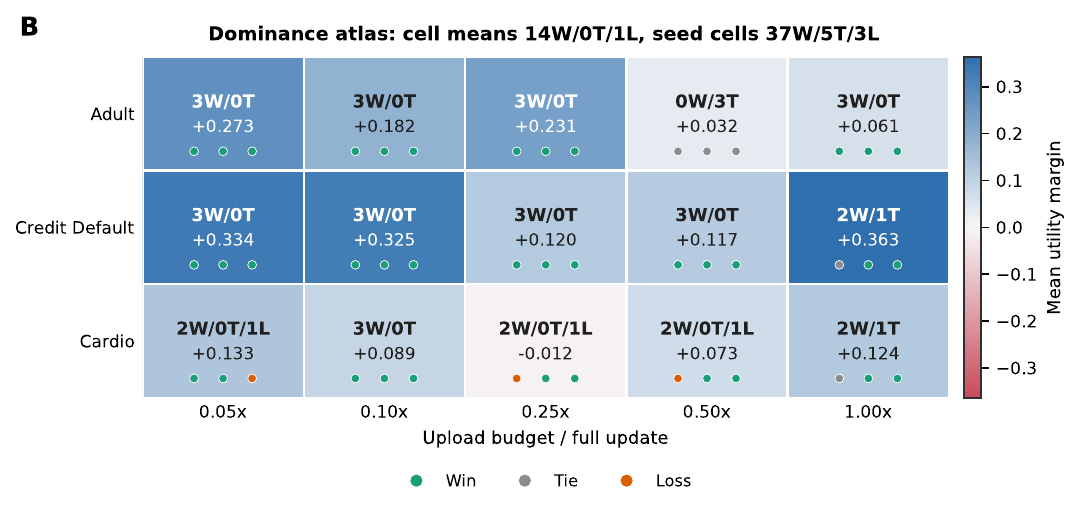}
    \caption{
    \textbf{Dominance atlas across datasets and budgets.}
    Cell colors denote FedLBAT's average utility margin against the strongest non-FedLBAT communication baseline.
    Cell text reports seed-level win/tie/loss counts.
    FedLBAT obtains positive mean margins in all $14$ dataset--budget cells.
    }
    \label{fig:dominance_atlas}
\end{figure}

Figure~\ref{fig:dominance_atlas} visualises the same conclusion as a dominance atlas.
Each cell corresponds to one dataset and one budget ratio.
The colour indicates the average utility margin of FedLBAT over competing communication-constrained baselines, and the text reports seed-level win/tie/loss counts.
The weakest region appears around $0.50\times$ on Adult, where FedLBAT ties the strongest peer across seeds; even there, the mean margin remains positive.
Overall, the dominance atlas confirms that LBAT's advantage is systematic rather than the result of isolated favourable cases.

\subsection{Interpretability of Adaptive Budget Allocation}

\begin{figure}[t]
    \centering
    \includegraphics[width=\columnwidth]{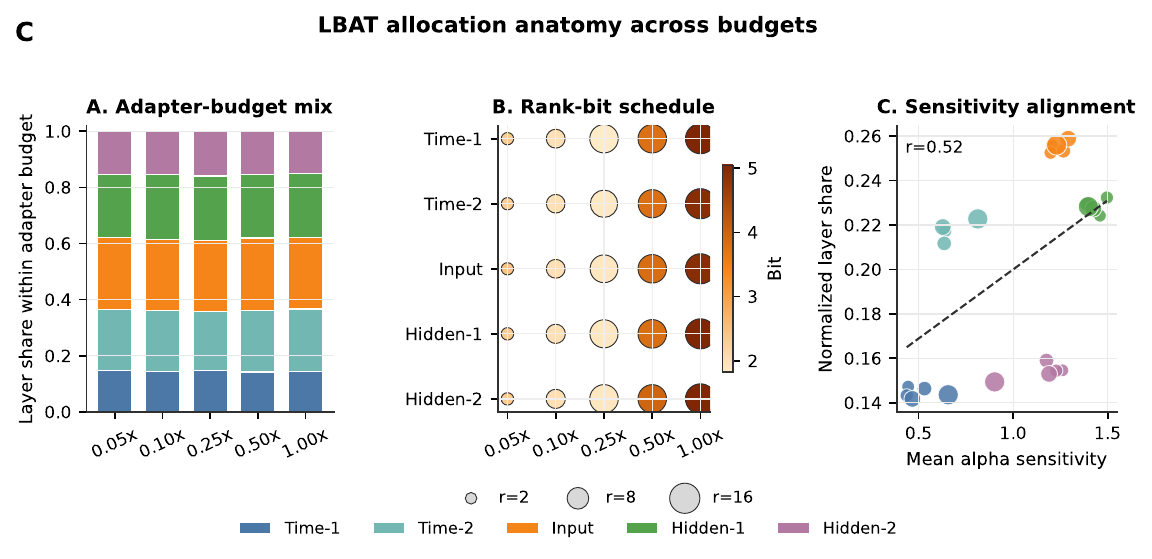}
    \caption{
    \textbf{LBAT allocation anatomy.}
    A: adapter-budget mix across blocks;
    B: selected rank--bit schedule;
    C: correlation between sensitivity and normalised layer share.
    LBAT allocates more communication to sensitive blocks and increases rank/bit adaptively as the budget grows.
    }
    \label{fig:policy_anatomy}
\end{figure}

Figure~\ref{fig:policy_anatomy} examines what LBAT actually learns.
Panel A shows the fraction of the adapter budget assigned to each block.
Panel B shows the rank--bit configurations selected under different budgets.
Panel C compares layer-wise sensitivity with the final normalised layer share.
The allocation is clearly non-uniform: LBAT does not simply spread the budget evenly across layers.
As the budget increases from $0.05\times$ to $1.00\times$, selected ranks and bit widths increase, but the relative budget shares remain layer-dependent.
Moreover, the positive correlation between sensitivity and allocated share ($r=0.52$) indicates that LBAT's local sensitivity proxy is aligned with its final transmission decisions.
This provides evidence that the algorithm learns a meaningful parameter-block value allocation, rather than merely acting as another uniform compression rule.

\subsection{Data Quality and Distribution Fidelity Diagnostics}

\begin{figure}[t]
    \centering
    \includegraphics[width=\columnwidth]{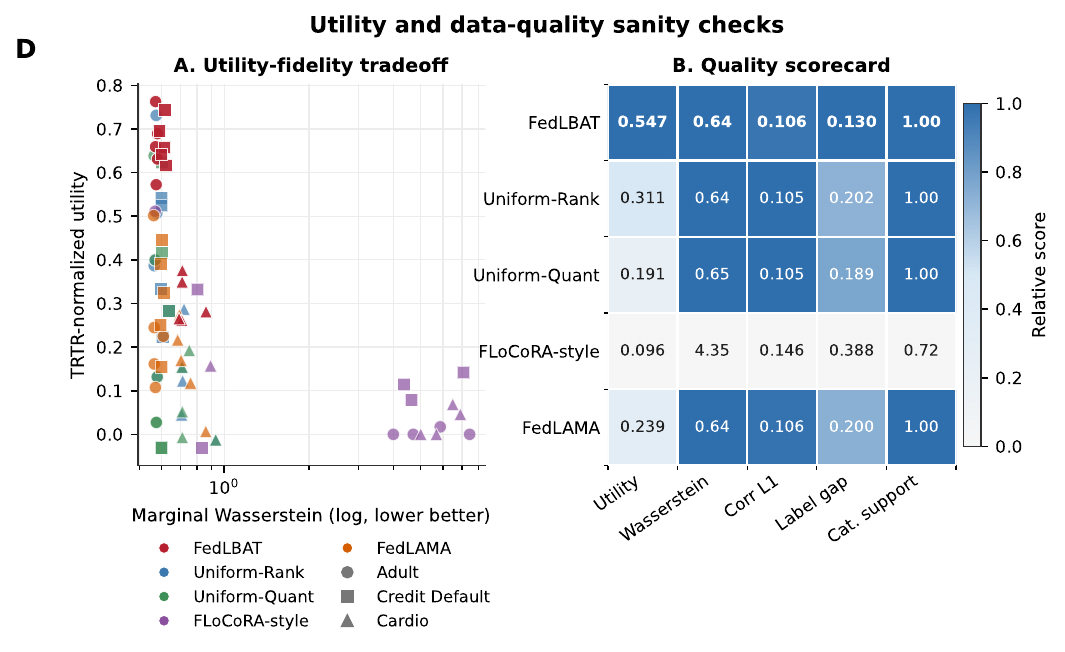}
    \caption{
    \textbf{Utility and data-quality diagnostics.}
    Left: utility--fidelity tradeoff.
    Right: relative quality scorecard.
    FedLBAT achieves the strongest utility while keeping distributional fidelity and categorical support competitive with or better than uniform baselines.
    }
    \label{fig:quality_sanity}
\end{figure}

High downstream utility is only meaningful if the generated tabular distribution remains faithful.
Figure~\ref{fig:quality_sanity} therefore reports utility--fidelity and data-quality diagnostics.
FedLBAT achieves the highest average normalised utility while maintaining competitive marginal Wasserstein distance, correlation error, label gap, and full categorical support.
In contrast, FLoCORA-style compression shows substantially weaker utility, higher Wasserstein distance, and lower categorical support.
This suggests that FedLBAT's utility gain does not simply come from overfitting the downstream classifier; it also preserves important tabular distributional properties.

\subsection{Generalization to Real Tabular Prediction}

Although our main evaluation focuses on communication-constrained tabular modelling, LBAT is designed as a generic layer-wise transmission policy.
We therefore also validate it on real tabular prediction.
Table~\ref{tab:track_summary} summarises results across the synthetic TSTR track and the real prediction track.
On the synthetic track, FedLBAT consistently improves over the strongest non-FedLBAT peer across low, medium, and full-budget regimes, with AUC gains of $+0.0778$, $+0.0489$, and $+0.0534$, respectively.
On the real prediction track, FedLBAT is essentially tied with the strongest peer baseline.
This indicates that LBAT does not degrade standard predictive federated learning, while its largest gains arise in the more difficult tabular modelling setting where layer values are more heterogeneous and uniform compression is more likely to waste budget.

\begin{table*}[t]
\centering
{%
\fontsize{9pt}{10pt}\selectfont
\setlength{\tabcolsep}{1mm}
\begin{tabular}{@{}llcccc@{}}
\toprule
Track
& Budget Group
& FedLBAT
& Best Peer
& $\Delta$ AUC
& Cells Won \\
\midrule
Syn.
& $0.05/0.10\times$
& \textbf{0.686 $\pm$ 0.081}
& 0.608 $\pm$ 0.075
& +0.0778
& 6/6 \\
Syn.
& $0.25/0.50\times$
& \textbf{0.685 $\pm$ 0.096}
& 0.636 $\pm$ 0.095
& +0.0489
& 6/6 \\
Syn.
& $1.00\times$
& \textbf{0.670 $\pm$ 0.080}
& 0.616 $\pm$ 0.118
& +0.0534
& 3/3 \\
\midrule
Real
& $0.05/0.10\times$
& 0.811 $\pm$ 0.068
& \textbf{0.811 $\pm$ 0.069}
& -0.0005
& 0/6 \\
Real
& $0.25/0.50\times$
& 0.810 $\pm$ 0.070
& \textbf{0.812 $\pm$ 0.068}
& -0.0021
& 2/6 \\
Real
& $1.00\times$
& 0.810 $\pm$ 0.073
& \textbf{0.811 $\pm$ 0.071}
& -0.0010
& 1/3 \\
\bottomrule
\end{tabular}
}

\caption{
Cross-track AUC summary.
Syn. denotes the synthetic TSTR track; Real denotes the real tabular prediction track.
Best Peer is the strongest non-FedLBAT communication-constrained baseline within each budget group.
Boldface indicates the higher mean AUC in each row, determined from the unrounded results.
}
\label{tab:track_summary}
\end{table*}

\begin{table*}[t]
\centering
{%
\fontsize{9pt}{10pt}\selectfont
\setlength{\tabcolsep}{1mm}
\begin{tabular}{@{}llcccccc@{}}
\toprule
Variant
& \shortstack{Ablated\\Component}
& Budget
& Adult
& Credit
& Cardio
& \shortstack{Avg.\\AUC}
& \shortstack{$\Delta$ vs.\\Full} \\
\midrule
\textbf{Full FedLBAT}
& None
& $0.05\times$
& \textbf{0.755 $\pm$ 0.056}
& \textbf{0.703 $\pm$ 0.037}
& \textbf{0.642 $\pm$ 0.050}
& \textbf{0.700}
& -- \\

w/o layer sensitivity
& Layer sensitivity
& $0.05\times$
& \textbf{0.755 $\pm$ 0.058}
& 0.655 $\pm$ 0.035
& 0.598 $\pm$ 0.048
& 0.669
& -0.031 \\

w/o drop-loss calib.
& Drop-loss calibration
& $0.05\times$
& \textbf{0.755 $\pm$ 0.052}
& 0.646 $\pm$ 0.041
& 0.601 $\pm$ 0.055
& 0.667
& -0.033 \\

w/o adaptive alloc.
& Budgeted allocation
& $0.05\times$
& 0.755 $\pm$ 0.061
& 0.540 $\pm$ 0.039
& 0.588 $\pm$ 0.047
& 0.628
& -0.072 \\

Random allocation
& Value-guided allocation
& $0.05\times$
& 0.689 $\pm$ 0.054
& 0.609 $\pm$ 0.034
& 0.568 $\pm$ 0.052
& 0.622
& -0.078 \\
\bottomrule
\end{tabular}
}

\caption{
Ablation study of FedLBAT under the strict $0.05\times$ communication budget.
Removing adaptive allocation, layer sensitivity, or drop-loss calibration consistently degrades performance, validating the value-guided dynamic-programming design.
Boldface marks the best mean in each column, with ties retained at the displayed precision.
}
\label{tab:ablation}
\end{table*}
Table~\ref{tab:ablation} confirms that both adaptive budget allocation and gradient-based layer sensitivity are crucial for FedLBAT's performance under extreme compression. Removing either component, or relying on random allocation, causes significant AUC drops across datasets, validating our value-guided dynamic programming design.

\section{Discussion}

Our results indicate that, under extreme budgets, the key issue is how scarce bytes are allocated rather than how aggressively every update is compressed. Uniform-Rank and Uniform-Quant impose the same configuration across blocks, while FLoCORA-style compression follows a fixed low-rank policy. FedLAMA adapts when layers are aggregated, but not how rank and precision are assigned within a round. FedLBAT instead compares candidate costs and distortions, then directs capacity toward blocks whose errors matter more. The moderate alignment between sensitivity and allocated share supports this mechanism, although it is not a causal guarantee.

The two evaluation tracks clarify when this allocation matters. Real prediction is close to saturation, so compressed updates retain enough discriminative information; FedLBAT is therefore largely tied with the strongest peers. Synthetic tabular modelling is less forgiving because it must preserve label structure, mixed-type dependencies, fidelity, and downstream predictability together. Uniform baselines retain fidelity but lower utility, whereas FLoCORA-style compression shows label and coverage failures. These patterns suggest that LBAT is useful when parameter-block values are heterogeneous. More broadly, LBAT complements quantisation, low-rank approximation, and temporal aggregation by treating them as candidate actions in a per-round allocation problem.

We note three limitations. First, expanding LBAT to sequential vision or foundation models requires exploration beyond tabular benchmarks. Second, our empirical sensitivity score is an efficient proxy rather than a formal utility guarantee, leaving room for stronger valuation methods. Third, although LBAT reduces transmitted information, it lacks formal privacy guarantees, making integration with differential privacy and client fairness important future directions.

\section{Conclusion}
We introduced LBAT to address communication bottlenecks in federated learning. Our work demonstrates that the primary challenge under extreme upload budgets is not merely how aggressively to compress an update. Instead, the core bottleneck lies in how to distribute limited communication resources across different parameter blocks. LBAT successfully transforms federated communication from a uniform compression paradigm into a budgeted value allocation problem. By utilising local sensitivity estimates and a dynamic programming allocator, LBAT prioritises bandwidth for layers that dominate downstream utility. Our empirical evaluation confirms that this adaptive allocation significantly improves utility while maintaining high data quality.

\bibliography{references}

\end{document}